\def\cvprPaperID{0001}
\def\confName{CVPR}
\def\confYear{2023}
\def\paperTitle{DPPD: Deformable Polar Polygon Object Detection}

\def\authorBlock{
    Yang Zheng \qquad
    Oles Andrienko \qquad
    Yonglei Zhao \qquad
    Minwoo Park \qquad
    Trung Pham \qquad \\
    NVIDIA \\
    {\tt\small \{yazheng, oandrienko, yongleiz, minwoop, trungp\}@nvidia.com}
}

\newif\ifreview 
\newif\ifarxiv \newcommand{\arxiv}{\arxivtrue}
\newif\ifcamera 
\newif\ifrebuttal 

\arxiv

\pdfoutput=1
\documentclass[10pt,twocolumn,letterpaper]{article}
\ifreview \usepackage[review]{cvpr} \fi
\ifarxiv \usepackage[pagenumbers]{cvpr} \fi
\ifrebuttal \usepackage[rebuttal]{cvpr} \fi
\ifcamera \usepackage{cvpr} \fi

\usepackage{graphicx}
\usepackage{amsmath}
\usepackage{amssymb}
\usepackage{booktabs}
\usepackage{color,soul}

\usepackage{times}
\usepackage{microtype}
\usepackage{epsfig}
\usepackage[table,xcdraw]{xcolor}
\usepackage{caption}
\usepackage{float}
\usepackage{placeins}
\usepackage{color, colortbl}
\usepackage{stfloats}
\usepackage{enumitem}
\usepackage{tabularx}
\usepackage{xstring}
\usepackage{multirow}
\usepackage{xspace}
\usepackage{url}
\usepackage{subcaption}
\usepackage{xcolor}
\usepackage{soul}
\usepackage[hang,flushmargin]{footmisc}
\usepackage[accsupp]{axessibility}  

\ifcamera \usepackage[accsupp]{axessibility} \fi

% ...

\ifarxiv  \fi

\newcommand{\R}[1]{{%
    \textbf{%
        \ifstrequal{#1}{1}{\textcolor{red}{R#1}}{%
        \ifstrequal{#1}{2}{\textcolor{blue}{R#1}}{%
        \ifstrequal{#1}{3}{\textcolor{magenta}{R#1}}{%
        \ifstrequal{#1}{4}{\textcolor{teal}{R#1}}{%
                           \textcolor{cyan}{R#1}%
        }}}}%
    }%
}}

\usepackage{xr-hyper}

\makeatletter
\newcommand*{\addFileDependency}[1]{
  \typeout{(#1)}
  \@addtofilelist{#1}
  \IfFileExists{#1}{}{\typeout{No file #1.}}
}

\makeatother

\usepackage[pagebackref,breaklinks,colorlinks]{hyperref}
\usepackage[capitalize]{cleveref}
\crefname{section}{Sec.}{Secs.}
\crefname{table}{Table}{Tables}
\crefname{figure}{Fig.}{Figs.}

\begin{document}
\title{\paperTitle}
\author{\authorBlock}
\maketitle

\begin{abstract}
Regular object detection methods output rectangle bounding boxes, which are unable to accurately describe the actual object shapes. Instance segmentation methods output pixel-level labels, which are computationally expensive for real-time applications. Therefore, a polygon representation is needed to achieve precise shape alignment, while retaining low computation cost.
We develop a novel \textit{Deformable Polar Polygon Object Detection} method (DPPD) to detect objects in polygon shapes. In particular, our network predicts, for each object, a sparse set of flexible vertices to construct the polygon, where each vertex is represented by a pair of angle and distance in the Polar coordinate system. To enable training, both ground truth and predicted polygons are densely resampled to have the same number of vertices with equal-spaced raypoints. The resampling operation is fully differentable, allowing gradient back-propagation. Sparse polygon predicton ensures high-speed runtime inference while dense resampling allows the network to learn object shapes with high precision. The polygon detection head is established on top of an anchor-free and NMS-free network architecture. \textit{DPPD} has been demonstrated successfully in various object detection tasks for autonomous driving such as traffic-sign, crosswalk, vehicle and pedestrian objects.

\end{abstract}
\section{Introduction}
\begin{figure}[t]
  \centering
  \includegraphics[width=0.9\linewidth]{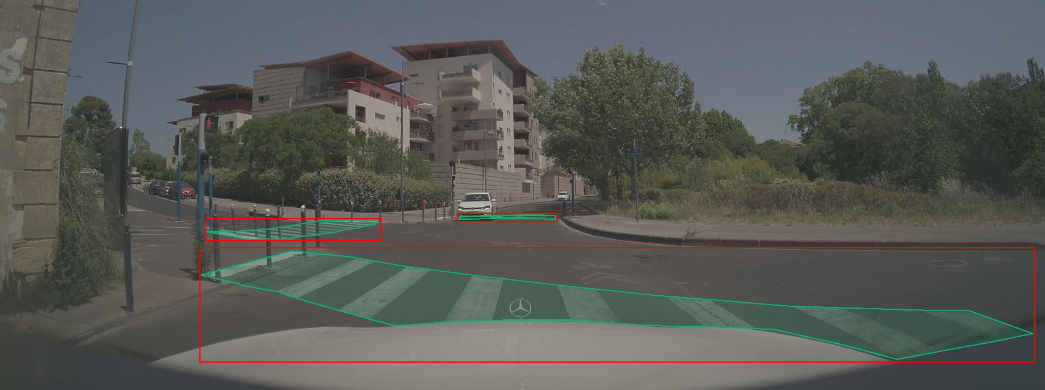}
  \caption{For crosswalk detection, polygon shapes are in green; bounding boxes are in red. It is clear that bounding boxes are unable to represent the crosswalk regions well as compared to polygons.}
  \label{fig:crosswalk_detection}
\end{figure}

Object detection, as one of the most popular computer vision tasks, typically predicts objects in rectangle bounding boxes. Boxes are able to describe locations and sizes, but not object shapes. Fig. \ref{fig:crosswalk_detection} shows an example of crosswalk detection for autonomous driving, where precise crosswalk regions are needed. This can be achieved by an instance segmentation method which outputs a pixel-wise mask per object. However, pixel-level post-processing is computationally expensive, thus not suitable for real-time applications.

\begin{figure}[t]
  \centering
  \includegraphics[width=0.9\linewidth]{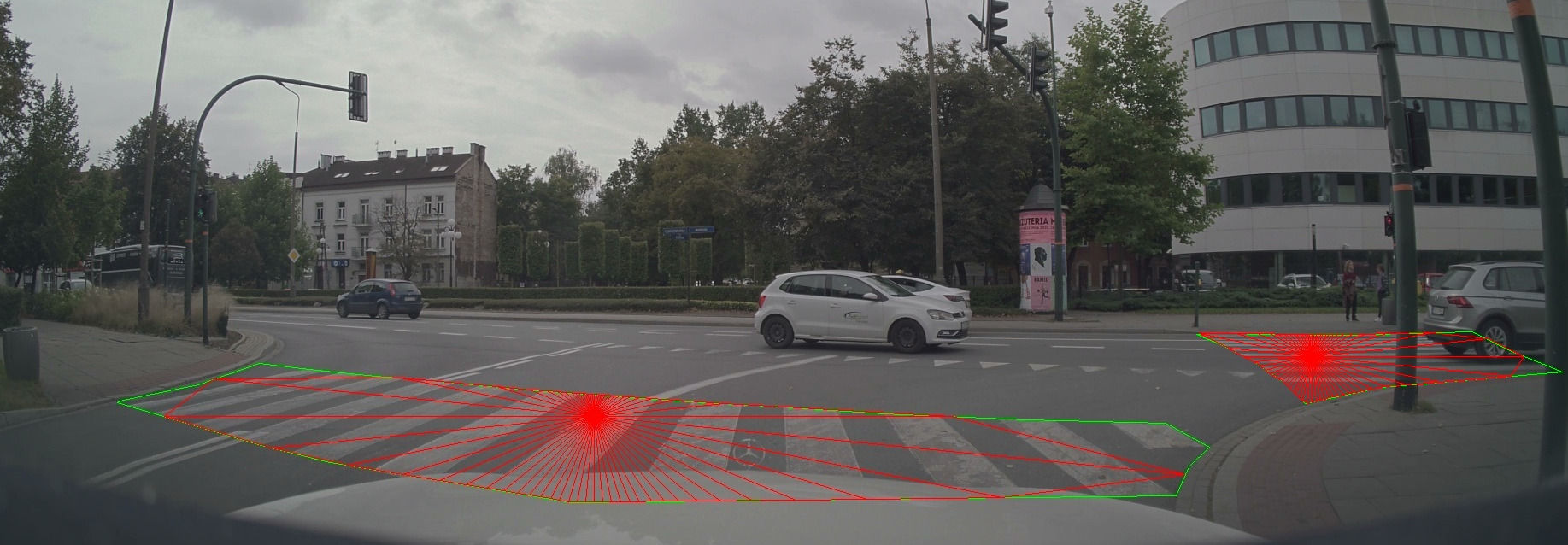}
  \caption{Example of approximating ground truth polygons (green) using polar polygons (red) with fixed angular bins. Notice that 64 rays is still insufficient to capture the actual crosswalk regions.}
  \label{fig:polarmask_rays}
\end{figure}

Alternatively, a method that detects objects as polygons is a better choice because 1) polygons can capture object shapes with high accuracy, and 2) a detection network can be run much faster than a segmentation network plus post-processing. Unfortunately, polygon detection is more complex than box detection in the variant numbers of vertices of arbitrary shapes. This creates difficulties when training a network to predict a fixed number of vertices. A common solution (\eg, as done in previous methods such as PolarMask \cite{xie2020polarmask}) is that ground truth polygons are represented in Polar coordinates and approximated by a vector of distance values and a (fixed) vector of evenly-spacing angular values. The task becomes training a network to regress, for each object, a radius vector, together with the predefined uniformly emitted rays decoded back to polygon. A clear limitation of this method is that the quality of ground truth labels (and thus the quality of prediction) is bounded by the number of rays. Fig \ref{fig:polarmask_rays} shows that even with 64 rays, crosswalk regions are not well captured. Increasing the numbers of rays will increase the computational cost significantly. 



\begin{figure*}
  \centering
  \includegraphics[width=0.75\linewidth]{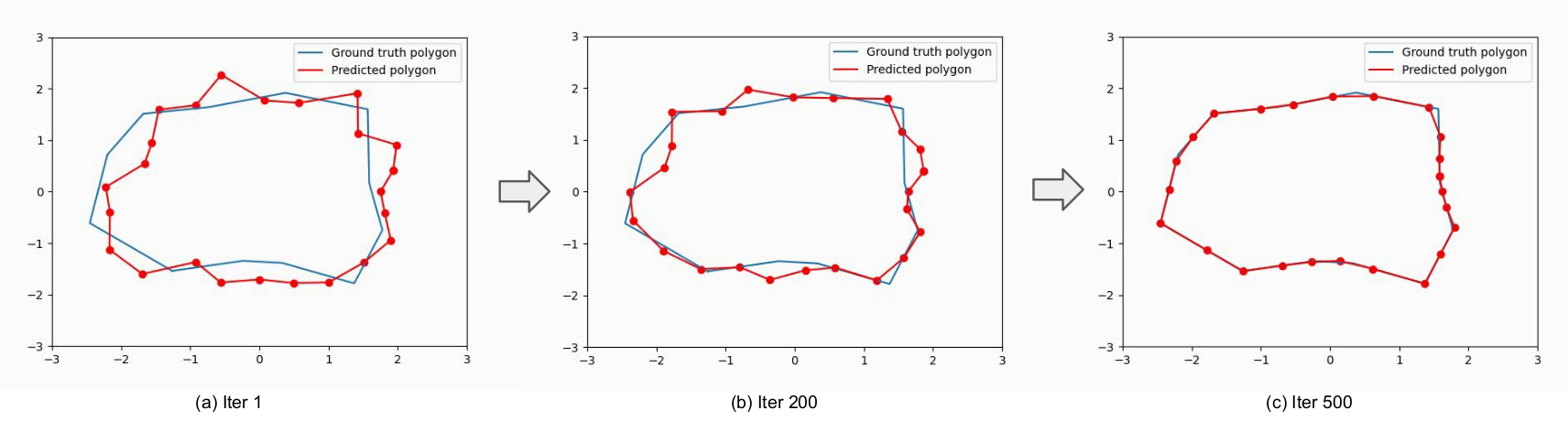}
  \caption{An illustration of the deformable polygon shape learning process. The target shape is in blue, and the predicted shape with 24 vertices is in red. From iteration step 1 to 200 to 500, the predicted shape is converging close to the target.}
  \label{fig:dppd_toy_example}
\end{figure*}

In this work, we propose a novel \textit{Deformable Polar Polygon Object Detection} method, namely \textit{DPPD}. Unlike PolarMask, our network predicts a small set of flexible polygon vertices directly, where each vertex has two degrees of freedom in Polar coordinates, \ie, radius and angle. Intuitively, starting from an initialized polygon, the network will deform it until it aligns well with the ground truth polygon (see Fig. \ref{fig:dppd_toy_example}). A question arises is how to compare and compute loss between ground truth and prediction where they are different in the number of vertices. Our idea is to densely resample both ground truth and prediction with a certain number of rays. We could resample as many rays as the memory allows. It's worth emphasizing that the resampling process only happens at the loss computation, thus does not affect the inference runtime, and that the resampling operation is differentiable, allowing gradient based optimization. 


Note that DPPD is designed for polygon shape regression, and it is applicable for any successful object detection architecture. In this work, we establish a high-speed, single-shot, anchor-free, NMS-free detection network structure based on set prediction \cite{zhang2019deep, rezatofighi2021learn} and build DPPD on top of it. This simple architecture escalates both training effectiveness and inference efficiency.


The main contributions of this work can be summarized as follows: 1) We propose a novel polygon detection method to detect objects with arbitrary shapes. Our method avoids a trade-off decision between network computation cost and polygon accuracy as in PolarMask; 2) a method to decode a regression vector into a valid polygon (\eg, vertices are in counter-clockwise order), a batching processing algorithm to resample sparse polygons to dense polygons with a minimum cost; 3) Inspired by the latest object detection development, we design a highly efficient single-stage, anchor-free and NMS-free polygon detection architecture; 4) Our proposed polygon detector surpasses previous polygon detection methods in both speed and accuracy when tested on autonomous driving perception tasks such as crosswalk, road sign, vehicle and pedestrian detection.

\label{sec:intro}

\section{Related Work}

\textbf{Object Detection.} Object detection has been evoluted from two-stage or one-stage anchor-based detectors \cite{ren2015faster, liu2016ssd, redmon2016you}, to anchor-free and NMS-free detectors \cite{duan2019centernet, zhou2019objects, tian2019fcos}, transformer-based detectors \cite{carion2020end}, and other variations \cite{lin2017focal, redmon2017yolo9000, redmon2018yolov3, huang2018yolo, bochkovskiy2020yolov4}.
In general, an object detector includes three major components: a backbone of series of convolutional blocks, a neck of multi-resolution feature pyramids \cite{lin2017feature}, and a detection head. The detection head is usually divided into a classification head and a bounding box regression head. Our detection method follows the anchor-free, NMS-free approach, but the box regression head is replaced by a polygon regression head to predict object locations and boundaries. 

\textbf{Instance Segmentation.} Instance segmentation produces pixel-level class-ids and object-ids. Mask R-CNN \cite{he2017mask} introduces a detect-then-segment approach to breakdown this problem into two sequential sub-tasks. To overcome the expensive two-stage processing, YOLOACT \cite{bolya2019yolact} and its extended methods \cite{chen2020blendmask, wang2020centermask, tian2020conditional} construct a parallel assembling framework, by generating a set of prototype masks and predicting per-instance mask coefficients.
However, regardless of model variations, the per-pixel segmentation mask output is always a huge burden for downstream real-time applications.

\textbf{Polygon Detection.} A series of work, such as ExtremeNet \cite{zhou2019bottom}, ESE-Seg \cite{xu2019explicit}, PolarMask \cite{xie2020polarmask}, and FourierNet \cite{riaz2021fouriernet}, try to parameterize the contour of an object mask into fixed-length coefficients, given different decomposition bases. These methods predict the center of each object and the contour shape with respect to that center. 
PolarMask is the one closest to our method, which is built on top of FCOS \cite{tian2019fcos} and utilizes depth-variant rays at constant angle intervals. 
Similarly, PolyYOLO \cite{hurtik2022poly} adopts the YOLOv3 \cite{redmon2018yolov3} architecture, and modifies the perpendicular grid into circular sectors to detect polar coordinates of polygon vertices. Each circular sector is responsible to produce 1 or 0 vertex.
The drawback of these methods is that the shape alignment quality is heavily bottlenecked by the pre-defined ray bases. Increasing the number of rays is possible to improve the quality, but meanwhile downgrading the speed performance. In contrast, our method is less dependent on the number of vertices. We found that as small as 12 vertices is sufficient to model variety of object shapes.

\textbf{Active Contour Model (ACM).} In the classical computer vision area, the active contour model \cite{kass1988snakes} has been used to describe object shape boundaries. The main idea is to minimize an dedicated internal and external energy functions. The external term is to control the contour shape fitting, and the internal term is to control the deformation continuity. Our DPPD is inspired by the ACM. The training loss jointly minimizes the shape fitting error between ground truth and prediction polygons and polygon smoothness.

\label{sec:related}

\section{Method}

\begin{figure*}
  \centering
  \includegraphics[width=0.8\linewidth]{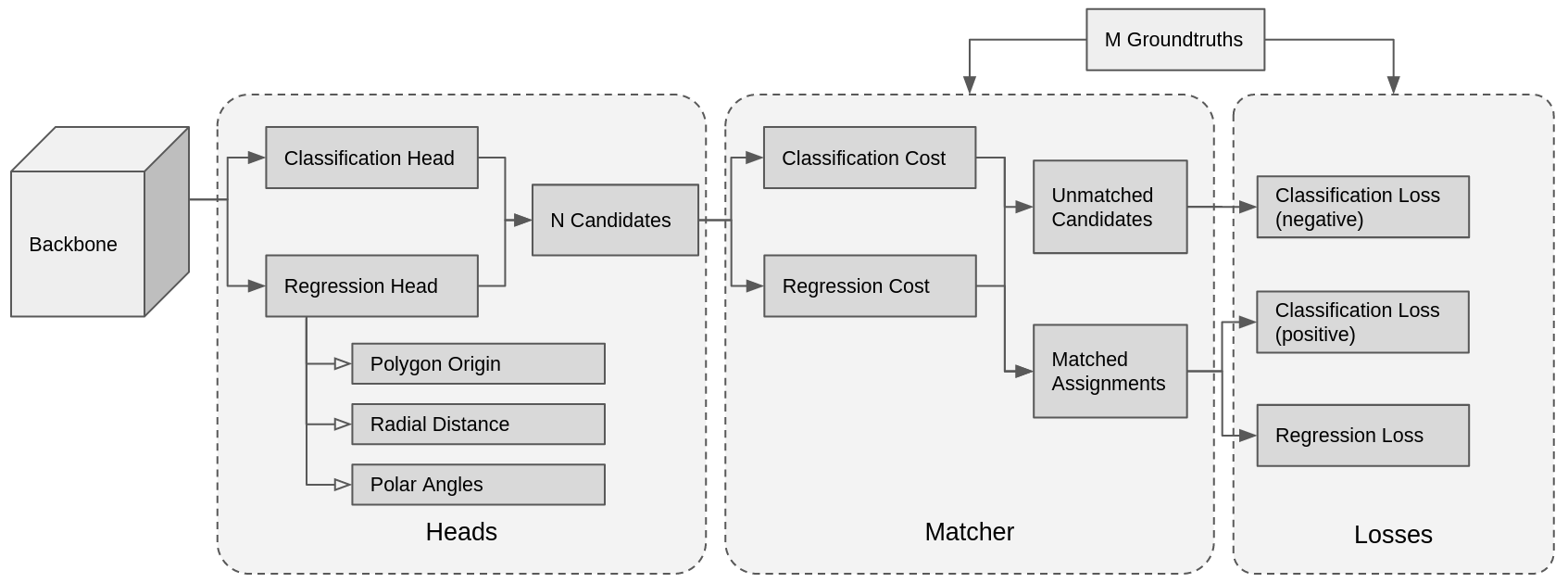}
  \caption{Overall network architecture and training pipeline. We establish a classification head and a regression head to generate $N$ candidates. The $N$ candidates are matched with $M$ groundtruths, resulting in $M$ pairs of prediction-target assignments. Based on the assignment, we compute positive classification and regression losses.}
  \label{fig:high_level_diagram}
\end{figure*}

In this section, we first introduce the overall set prediction network architecture. We then describe the polygon detection head. And lastly, we discuss the training strategies.

\subsection{Object Detection as Set Prediction}
We adopt an anchor-free and NMS-free set prediction approach for our DPPD object detector due to its simplicity  and efficiency. The network predicts a set of $N$ candidates ($N \gg M$ number of ground truth objects). The $N$ candidates are matched with $M$ ground truth labels using a Hungarian matching algorithm, resulting in $M$ pairs of prediction-target assignments. Classification and regression losses from these matches are computed to supervise the training. For unmatched candidates, only classification loss is computed.

\cref{fig:high_level_diagram} depicts the high-level network architecture and training pipeline. Followed by the network backbone and feature pyramid is a classification head and a regression head. 
The regression head predicts polygon origins (\ie, object center) and vertices (\ie, radial distance and polar angles). One grid cell in the feature map is responsible for detecting one polygon candidate.

\subsection{Polygon Regression}
\subsubsection{Polar Representation}
In the polar coordinates, each polygon is represented as one origin (object center) and $k$ pairs of radial distances and polar angles. Distances and angles are defined w.r.t the object center. The network will output a $(2+2*k)$-vector, where $2$ values are for the polygon origin and $2*k$ values are for $k$ vertices. 
The contouring vertices, in the polar representation, are convenient to be organized a clockwise or counterclockwise order.

\subsubsection{Polygon Decoding}
The decoding process parses a regression output vector $[f_{0}, f_{1}, ..., f_{2*k+2}]$ to the corresponding polygon origin coordinates, radial distances, and polar angles, denoting as $[o_{x}, o_{y}, r_{0}, ..., r_{k-1}, a_{0}, ..., a_{k-1}]$.

\textbf{Polygon origin} 
In the fully convolutional set prediction framework, every grid cell at the feature map yields a candidate. To get the accurate location, we predict offsets w.r.t. the grid cell position. Formally, the polygon origin is decoded as:
\begin{equation}
  \begin{aligned}
    	\begin{cases}
  	  o_{x} = g_{x} + s_{x} * \sigma(f_{0}) \\
  	  o_{y} = g_{y} + s_{y} * \sigma(f_{1})
  	\end{cases}
  \end{aligned}  
  \label{eq:3}
\end{equation}
where $(o_{x}, o_{y})$ denote the polygon origin coordinates; $(g_{x}, g_{y})$ denote the grid cell coordinates; $(s_{x}, s_{y})$ denote the grid cell size; $\sigma$ is a sigmoid activation function.

\textbf{Radial distances} 
The next $k$ regression outputs $(f_{2}, ..., f_{k+2})$ are dedicated for $k$ radial distances. The decoding function is:
\begin{equation}
  \begin{aligned}
        r_{i} = \mu * e^{f_{i}}, i \in [2, k+2]
  \end{aligned}
  \label{eq:4}
\end{equation}
where $\mu$ is a prior knowledge of the radius scale. We apply an exponential activation to ensure the decoded radius is always positive.

\textbf{Polar angles}
The last $k$ output channels $(f_{k+2}, ..., f_{2*k+2})$ are responsible for the $k$ polar angles. We predict angle deltas between adjacent vertices, and then decode using cumulative sum before normalizing them into $[0, 2\pi]$ range:
\begin{equation}
  \begin{aligned}
  	a_{i} = 2 \pi * \dfrac{\sum_{j=k+2}^{i}e^{f_{j}}}{\sum_{j=k+2}^{2*k+2}e^{f_{j}}},  i \in [k+2, 2*k+2]
  \end{aligned}
  \label{eq:5}
\end{equation}

It can be seen clearly that unlike the previous methods such as PolarMask \cite{xie2020polarmask}, which predefined polar angles (e.g., by shooting uniform rays from 0 to 360 degrees), our method predicts both polar angles and radial distances. This allows the network to deform the initial polygons as much as needed to match the ground truth polygons. In contrast, previous methods find difficulties to fit well the ground truth shapes unless a dense number of rays (\eg, 360) is used.


Note that the angle decoder also differentiates DPPD against Poly-YOLO \cite{hurtik2022poly}. Poly-YOLO splits the polar coordinates into circular sectors where each sector is responsible for either 1 (exist) or 0 (non-exist) vertex. This design is unable to discriminate clustered vertices that fall into the same sector. Instead, DPPD predicts angle deltas, which could be small and large to handle arbitrary intervals.


\subsection{Training}

\subsubsection{Ground Truths}
Objects are often annotated using polygons. However, there is no consistent way to enforce all the objects having the same number of vertices and vertex distributions along the boundaries. 
Therefore, it is less reasonable to train a network to predict polygons with a fixed number of vertices and by directly comparing ground truth and predicted vertices as often done in the box regression problem. 

Instead, we propose a simple method to tackle this issue, in which both ground truth and prediction polygons are resampled to have the same number of points, namely \textit{raypoints}.
Bear in mind that this resampling process only happens during training, thus does not hinder inference latency. This simple idea seems overlooked in the past mainly because the resampling operation might be not efficient, and not differentiable for SGD based training. Below we will describe two resampling methods which are both efficient and differentiable.  

\subsubsection{Polygon Resampling}

Given a polygon with $k$ vertices, we want to upsample it with $m$ ($m > k$) raypoints along equally distributed polar angles. This resampling process is a geometry problem, \ie, to find intersections between $k$ boundary segments and $m$ rays. To simplify the computation, we assume the polygon is translated to its origin at $(0, 0)$. In polar coordinates, since rays are uniformly emitted with the same angle interval, the resampling output only consists a $m$ dimensional radial distance vector $[r_{0}, r_{1}, ..., r_{m-1}]$. We provide two approaches to tackle this problem, triangle approach and vector approach. 


\begin{figure}
  \centering
  \begin{subfigure}{0.45\linewidth}
  \includegraphics[width=1.0\linewidth]{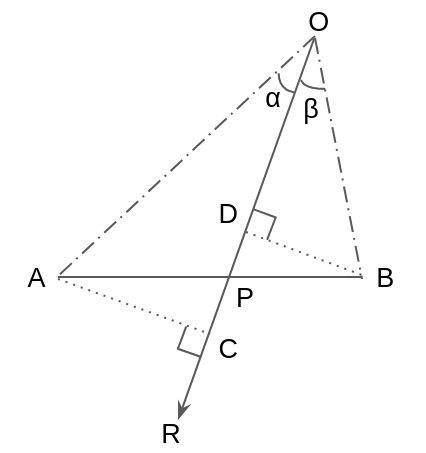}
  \caption{Triangle approach.}
    \label{fig:5-a}
  \end{subfigure}
  \hfill
  \begin{subfigure}{0.45\linewidth}
	\includegraphics[width=1.0\linewidth]{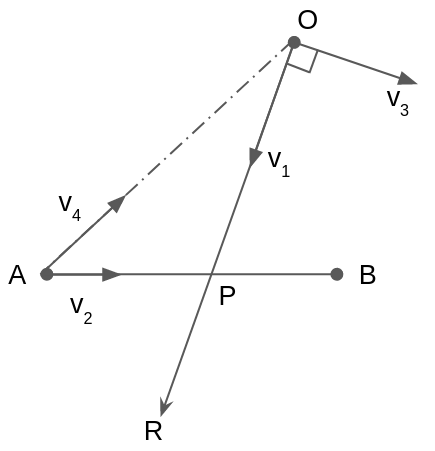}
    \caption{Vector approach.}
    \label{fig:5-b}
  \end{subfigure}
  \caption{$O$ is the polygon center, $A$ and $B$ are two adjacent vertices, $\vec{OR}$ is one ray emitted from $O$. The goal is to find the intersection point $P$ between the segment $AB$ and ray $\vec{OR}$. Refer to \cref{sec:resampling} for detailed derivation.}
  \label{fig:5}
\end{figure}

\textbf{Triangle Approach.}
In \cref{fig:5-a}, let $A$ and $B$ be two adjacent vertices and $O$ be the origin. The task is to find the intersection point $P$ between segment $AB$ and ray $\vec{OR}$. The norm $|\cdot|$ notation is used for the segment length. Deriving from the triangle similarity between $\triangle ACP$ and $\triangle BDP$:

\begin{equation}
  \begin{aligned}
  	w = \dfrac{|AP|}{|BP|} = \dfrac{|AC|}{|BD|} = \dfrac{|OA|\sin(\alpha)}{|OB|\sin(\beta)}.
  \end{aligned}
  \label{eq:6}
\end{equation}

we compute a length ratio $w$ between $|AP|$ and $|AB|$. Then the point $P$ coordinates are calculated as:

\begin{equation}
  \begin{aligned}
  	(P_x, P_y) = (\dfrac{A_x + w B_x}{1+w}, \dfrac{A_y + w B_y}{1+w}).
  \end{aligned}
  \label{eq:7}
\end{equation}

The coordinates of $O$, $A$, $B$ and their segment lengths are known from the decoding outputs. $\vec{OR}$ is one of the $m$ equal-spaced resampling rays. Given a list of polygon vertices with their sorted polar angles, we can easily find, for each ray, a pair of neighboring vertices $A$, $B$ via linear search so that the ray locates between $\vec{OA}$ and $\vec{OB}$.

\textbf{Vector Approach.} 
In \cref{fig:5-b}, for each known segment $AB$ and ray $\vec{OR}$, let define the following 4 vectors:
\begin{equation}
  \begin{aligned}
	\begin{cases}
  	  \vec{v_1} = (\vec{OR}_x, \vec{OR}_y) \\
  	  \vec{v_2} = (B_x - A_x, B_y - A_y) \\
            \vec{v_3} = (-\vec{OR}_y, \vec{OR}_x) \\
  	  \vec{v_4} = (O_x - A_x, O_y - A_y)
  	\end{cases}
  \end{aligned}
  \label{eq:unit_vectors}
\end{equation}

Since $P$ is the intersection point by $\vec{OR}$ and $AB$, we formulate the vector representation of $\vec{OP}$ and $\vec{AP}$:
\begin{equation}
  \begin{aligned}
	\begin{cases}
  	  \vec{OP} = O + t_{1}\vec{v_{1}},  t_{1} \in [0, \infty) \\
  	  \vec{AP} = A + t_{2}\vec{v_{2}},  t_{2} \in [0, 1]
  	\end{cases}
  \end{aligned}
  \label{eq:8}
\end{equation}

where $t_{1}$ and $t_{2}$ are the fractional scale of units along $\vec{OR}$ and $\vec{AB}$. Finding the intersection is to solve $t_{1}$ and $t_{2}$ from \cref{eq:8}. We constrain $t_{1} \in [0, \infty)$ to ensure $P$ is along the positive direction of ray $\vec{OR}$, and $t_{2} \in [0, 1]$ to ensure $P$ is inside segment $AB$. The mathematical solution is:

\begin{equation}
  \begin{aligned}
	\begin{cases}
  	  t_{1} = \dfrac{|\vec{v_{2}} \times \vec{v_{4}}|}{\vec{v_{2}} \cdot \vec{v_{3}}} \\
  	  t_{2} = \dfrac{\vec{v_{4}} \cdot \vec{v_{3}}}{\vec{v_{2}} \cdot \vec{v_{3}}}
  	\end{cases}
  \end{aligned}
  \label{eq:9}
\end{equation}

where the operator $\cdot$ is the dot product and $\times$ is the cross product. Suppose $\vec{v_{1}}$ is decomposed into $[v_{1x}, v_{1y}]$, combining \cref{eq:8}, the coordinates of P is written as:

\begin{equation}
  \begin{aligned}
  	(P_x, P_y) = (O_{x} + t_{1} v_{1x}, O_{y} + t_{1} v_{1y}).
  \end{aligned}
  \label{eq:10}
\end{equation}

Note that the triangle approach assumes vertices have been ordered ascendingly. For predictions, we decode polar angles in counter-clockwise order, which naturally satisfies the assumption. However, for ground truth encoding, this is not guaranteed for concave shapes. Therefore, the triangle approach is only used for prediction decoding. 
On the other hand, the vector approach has no requirements on the vertex order, and it works for both convex and concave shapes. However, the cross-product calculation expense much memory in the practical implementation, therefor vector approach is only used for ground truth encoding.


\label{sec:resampling}

\subsubsection{Polygon Regression Losses}

A common way to measure the loss between two shapes is based on their intersection-over-union (IoU). For general polygons, there is no exact closed-form solution. Fortunately, with the polar representation and the above resampling strategy, computing losses between ground truth and prediction polygons becomes easier. Formally, the polygon shape regression loss $\mathcal{L}_{reg}$ is a weighted sum of three components: polygon origin loss $\mathcal{L}_{o}$, polar IoU loss $\mathcal{L}_{iou}$ and internal smoothness loss $\mathcal{L}_{sm}$:
\begin{equation}
  \begin{aligned}
  	\mathcal{L}_{reg} = w_{1}\mathcal{L}_{o} + w_{2}\mathcal{L}_{iou} + w_{3}\mathcal{L}_{sm}
  \end{aligned}
  \label{eq:10}
\end{equation}

\textbf{Polygon origin loss.} 
The origin loss $\mathcal{L}_{o}$ measures the difference between two polygons' centers. We employ smooth-$l_1$ loss ($l^s_1$) for the absolute distance error. Also the losses are normalized by the ground truth size. Formally, let $[\hat{x}_o, \hat{y}_o]$ be the ground truth center, $[\hat{w}, \hat{h}]$ be the polygon width and height, $[{x}_o, {y}_o]$ be the prediction center, the origin loss is computed as:
\begin{equation}
  \begin{aligned}
  	\mathcal{L}_{o} = \dfrac{l^s_1(o_{x}, \hat{o}_{x})}{\hat{w}} + \dfrac{l^s_1(o_{y}, \hat{o}_{y})}{\hat{h}}
  \end{aligned}
  \label{eq:11}
\end{equation}


\textbf{Polar IoU loss.} 
The polar IoU loss $\mathcal{L}_{iou}$ measures shape difference between the two polygons regardless of their locations. Let $[\hat{r}_0, \hat{r}_1, ..., \hat{r}_{m-1}]$ and $[r_0, r_1, ..., r_{m-1}]$ be the ground truth and prediction radial distances respectively, the polar IoU loss is computed as:
\begin{equation}
  \begin{aligned}
  	\mathcal{L}_{iou} = \log \dfrac{\sum_{i=0}^{m-1}\max(r_{i}, \hat{r_{i}})}{\sum_{i=0}^{m-1}\min(r_{i}, \hat{r}_{i})}
  \end{aligned}
  \label{eq:12}
\end{equation}

\textbf{Smoothness loss.} 
The smoothness loss is added to reduce the shape oscillation, similar to the internal energy of the classical active contour model \cite{kass1988snakes}. Let $[r_{0}, r_{1}, ..., r_{m-1}]$ be the prediction radial distances, $d^1r$ and $d^2r$ be the 1st and 2nd ordered differences, the smoothness loss is computed as:
\begin{equation}
  \begin{aligned}
  	\mathcal{L}_{sm} = \dfrac{\sum_{i=1}^{m-1}d^1r_i}{m-1} + \dfrac{\sum_{i=1}^{m-1}d^2r_i}{m-1}
  \end{aligned}
  \label{eq:13}
\end{equation}
\label{sec:method}

\section{Experiment}

We conduct experiments using two datasets: our in-house autonomous driving dataset, and the public Cityscapes \cite{cordts2015cityscapes} dataset. 
Since one major application of the polygon detector is for autonomous driving perception, we use our in-house dataset for the primary investigation. We focus on road-sign and crosswalk detections where accurate object boundaries are important for subsequent tasks such as localization, sign recognition. We examine the effectiveness of polygon detectors against bounding box detectors, and thoroughly compare DPPD with PolarMask \cite{xie2020polarmask} in terms of both accuracy and speed performance. 
To further demonstrate the generic effectiveness, we benchmark DPPD results on Cityscapes, which covers more dynamic instances like vehicles and pedestrians. We also explore ablation studies for the model design. 

\subsection{DPPD on Internal Datasets}

The in-house dataset statistics for road sign and crosswalk are summarized in \cref{tab:datasets}, where all labels are given as polygon vertices. 
In this experiment, we establish DPPD polygon head on top of a FPN augmented AlexNet network backbone. The number of channels for different CNN blocks are 32, 128, 256, 512 respectively. The input image size is $960 \times 480$. The feature maps at stride 8 and 32 are used for the crosswalk and road-sign tasks separately.


\begin{table}\small
  \caption{Number of instances for traffic-sign and crosswalk polygon detection datasets.}
  \centering
  \begin{tabular}{c c c}
    \toprule
    Task & Training & Evaluation \\
    \midrule
    Road sign & 377.4K & 39.6K \\
    Crosswalk & 260.2K & 36.1K \\
    \bottomrule
  \end{tabular}
  \label{tab:datasets}
\end{table}

\subsubsection{Polygon vs Bounding Box}

\begin{figure*}
    \centering
    \begin{subfigure}[b]{0.45\textwidth}   
        \centering 
        \includegraphics[width=\textwidth]{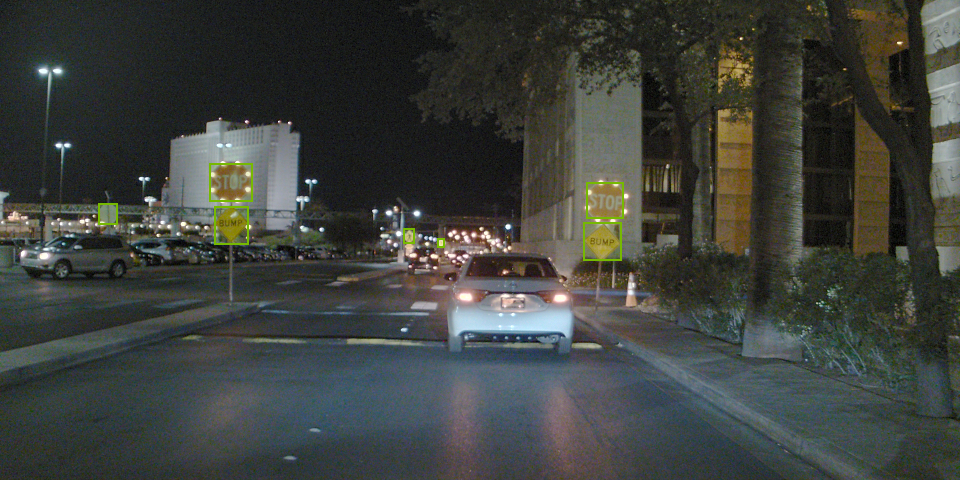}
    \end{subfigure}
    \begin{subfigure}[b]{0.45\textwidth}   
        \centering 
        \includegraphics[width=\textwidth]{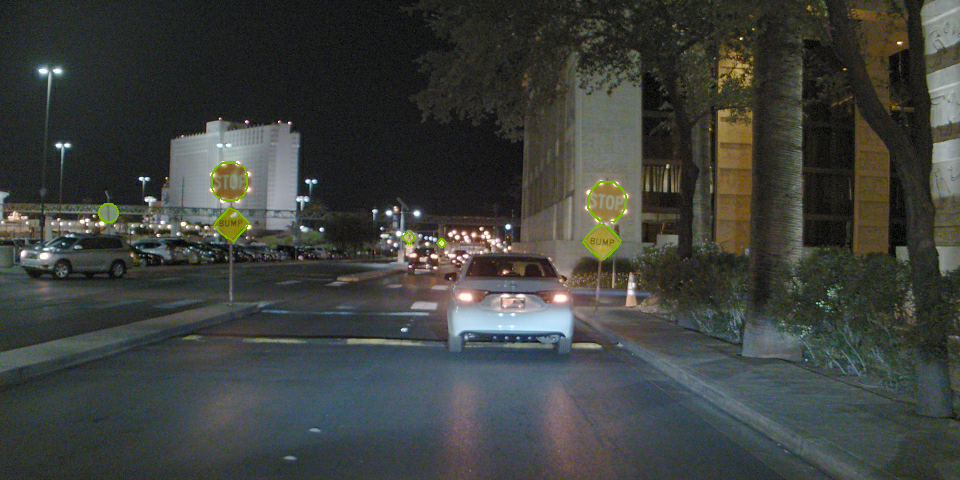}
    \end{subfigure}
    \caption{Two examples to visualize bounding box (left) and DPPD polygon (right) detections on traffic-signs. The polygon detector is better to capture the object shape.}
    \label{fig:ts-vis}
\end{figure*}


We first demonstrate the polygon detector is substitute for the bounding box detector. Typically, road signs are detected as bounding boxes. To swap it with a polygon model, we ensure that the new polygon model has non-negative effects on both detection accuracy and inference latency.

In this experiment, we set the number of predictable vertices $k=12$, and the number of resampling rays $m=180$. To compare the polygon against the box model, we convert polygons to their envelop bounding boxes, and evaluate both detectors based on the box metrics. Results are listed in \cref{tab:ts-res}. It shows that the polygon model is on-par with the box model with a very slight regression, although the polygon model was not trained to detect boxes. Nevertheless, the polygon model produces tighter alignments to the objects, which are qualitatively visualized in \cref{fig:ts-vis}.



The speed performance is device-dependent. We measure the runtime inference latency (in $ms$) when the model is deployed to TensorRT and run in FP16 precision on two hardware platforms: 2080 Titan GPU, and NVIDIA DRIVE Orin chip. The polygon detector runs $0.09ms$ slower on GPU, while $0.13ms$ faster on the chip. We consider the inference time is comparable on both platforms. This is expected because only the last regression layer is changed, whereas the change of regression channels is minor w.r.t. the total number of model parameters.

\begin{table}\small
     \caption{Traffic-sign results. Box vs. polygon detector. Accuracy metrics precision ($P$), recall ($R$), F1-score ($F1$) are in percentages. The speed is measured as latency times on Titan GPU and Orin Chip devices, in $ms$.}
        \centering
  		\begin{tabular}{c | c c  c | c c}
    			\toprule
    			Detector	& $P$ & $R$ & $F1$ & GPU & Chip \\
    			\midrule
    			BBox		& 74.91	& 58.68	& 65.81 &  1.54 & 3.18 \\
    			Polygon	& 74.54	& 57.31	& 64.80 & 1.63 & 3.05 \\
    			\bottomrule
  		\end{tabular}
     \label{tab:ts-res}
\end{table}


\subsubsection{DPPD vs PolarMask}

\begin{figure*}
    \centering
    \begin{subfigure}[b]{0.45\textwidth}
        \centering
        \includegraphics[width=\textwidth]{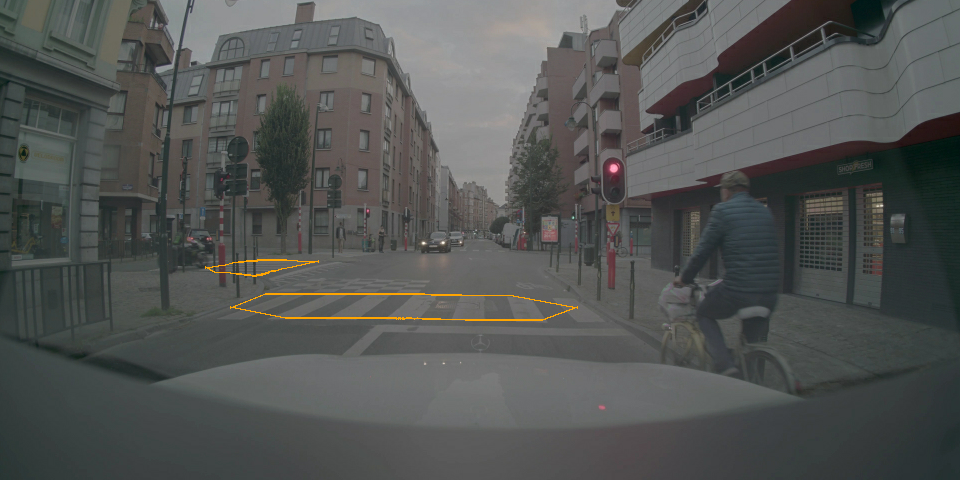}
    \end{subfigure}
    \begin{subfigure}[b]{0.45\textwidth}  
        \centering 
        \includegraphics[width=\textwidth]{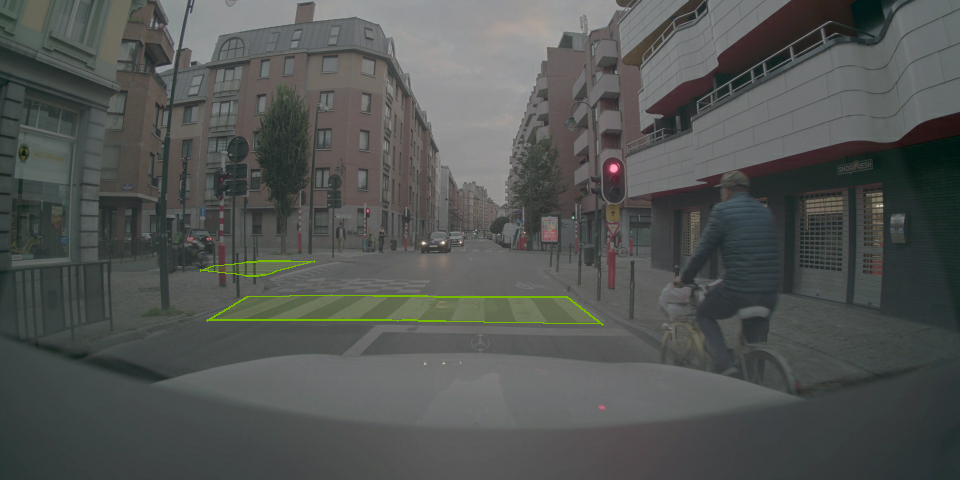}
    \end{subfigure}
    \caption{Two examples to visualize 64-vertex PolarMask (left) and 36-vertex DPPD (right) results on crosswalks. DPPD expenses less vertices but achieves tighter alignment with the groundtruth.}
    \label{fig:cw-vis}
\end{figure*}

Next we emphasize the benefits of DPPD against the state-of-the-art polygon detectors. We pick PolarMask as our competitor, and use the crosswalk detection task for this experiment. In DPPD, we set the number of prediction vertices $k=36$, and the number of resampling rays $m=360$. In PolarMask, we experimented different number of rays, \ie, $36$ and $64$. Since crosswalk shapes are more complex, the matching criteria is based on polygon-to-polygon IoU directly. Metrics results and visualizations are shown in \cref{tab:cw-res} and \cref{fig:cw-vis} respectively.

In terms of accuracy, DPPD is superior than PolarMask in all metrics, even the number of predictable vertices is less in DPPD ($36$) than PolarMask ($64$). The first reason is that PolarMask ground truth are approximated radius length along pre-defined rays, which are not guaranteed to reach "real" vertices (\eg, \cref{fig:polarmask_rays}). The second reason is that DPPD predicts, for each vertex, both radial distance and polar angle, which allows greater capability to achieve high-quality shape regression.

In terms of the speed, DPPD also costs less inference time than PolarMask on both platforms. We claim it is benefitted from the design of separate training and inference features. Dense polygons for training facilitates the accuracy, and sparse polygon for inference boosts the runtime speed. 

\begin{table}\small
    \caption{Crosswalk results. PolarMask vs. DPPD. The accuracy is evaluated based on polygon IoU directly. The speed is measured as latency times on GPU and Chip devices, in $ms$.}
    \centering
  	\begin{tabular}{c | c c c | c c}
    		\toprule
    		Detector	& $P$ & $R$ & $F1$ & GPU & Chip \\
    		\midrule
		    PolarMask-36	& 61.58	& 38.00 & 47.00 & - 	& - 		\\
		    PolarMask-64	& 66.72	& 45.46 & 54.08 & 1.91 	& 3.24 	\\
    		DPPD-36      	    & 77.27	& 50.46 & 61.05	& 1.28 	& 2.68	\\
    		\bottomrule
  	\end{tabular}
     \label{tab:cw-res}
\end{table}

\subsection{DPPD on Public Dataset}

\begin{figure*}
    \centering
    \begin{subfigure}[b]{0.45\textwidth}   
        \centering 
        \includegraphics[width=\textwidth]{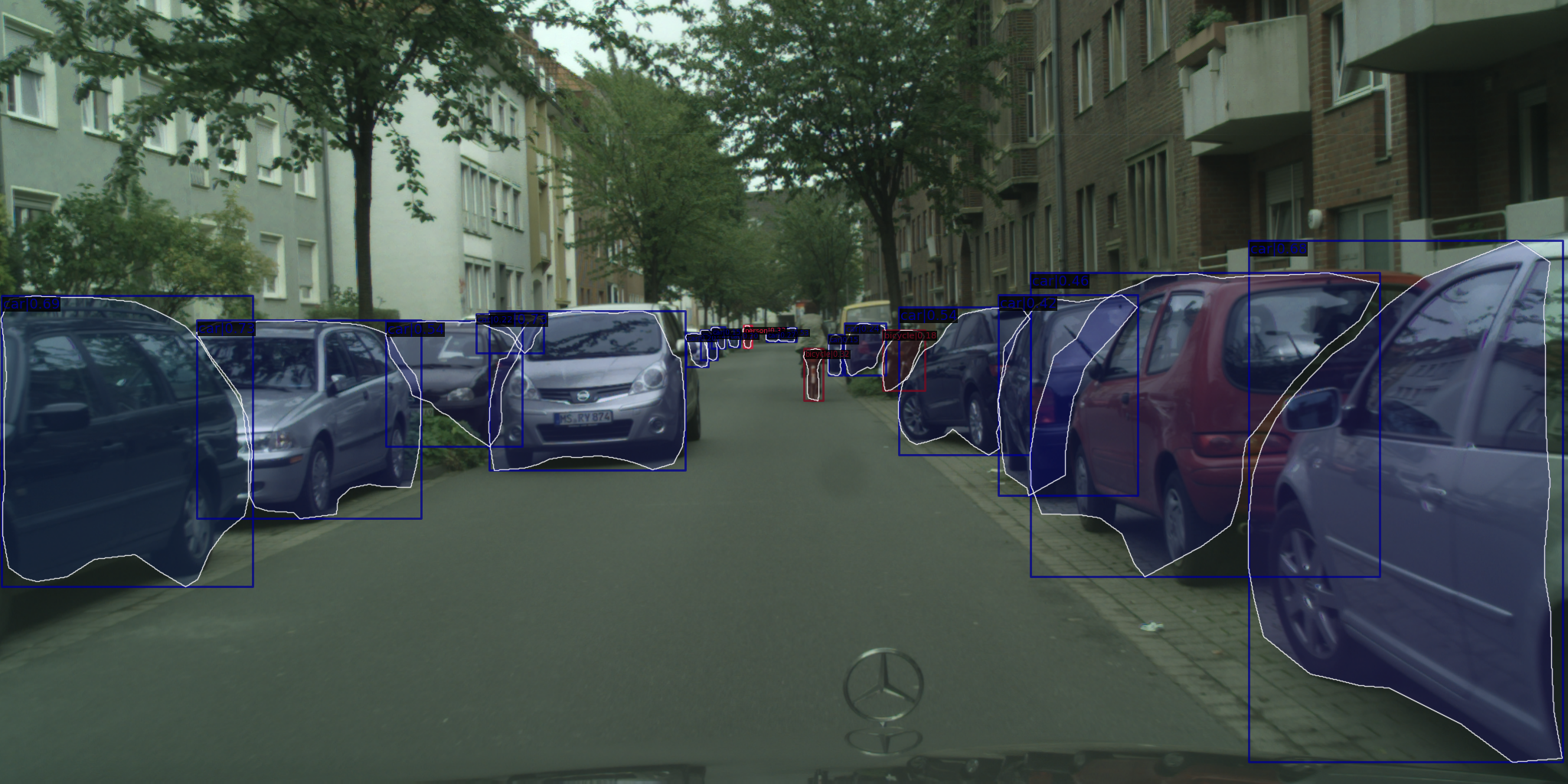}
    \end{subfigure}
    \begin{subfigure}[b]{0.45\textwidth}   
        \centering 
        \includegraphics[width=\textwidth]{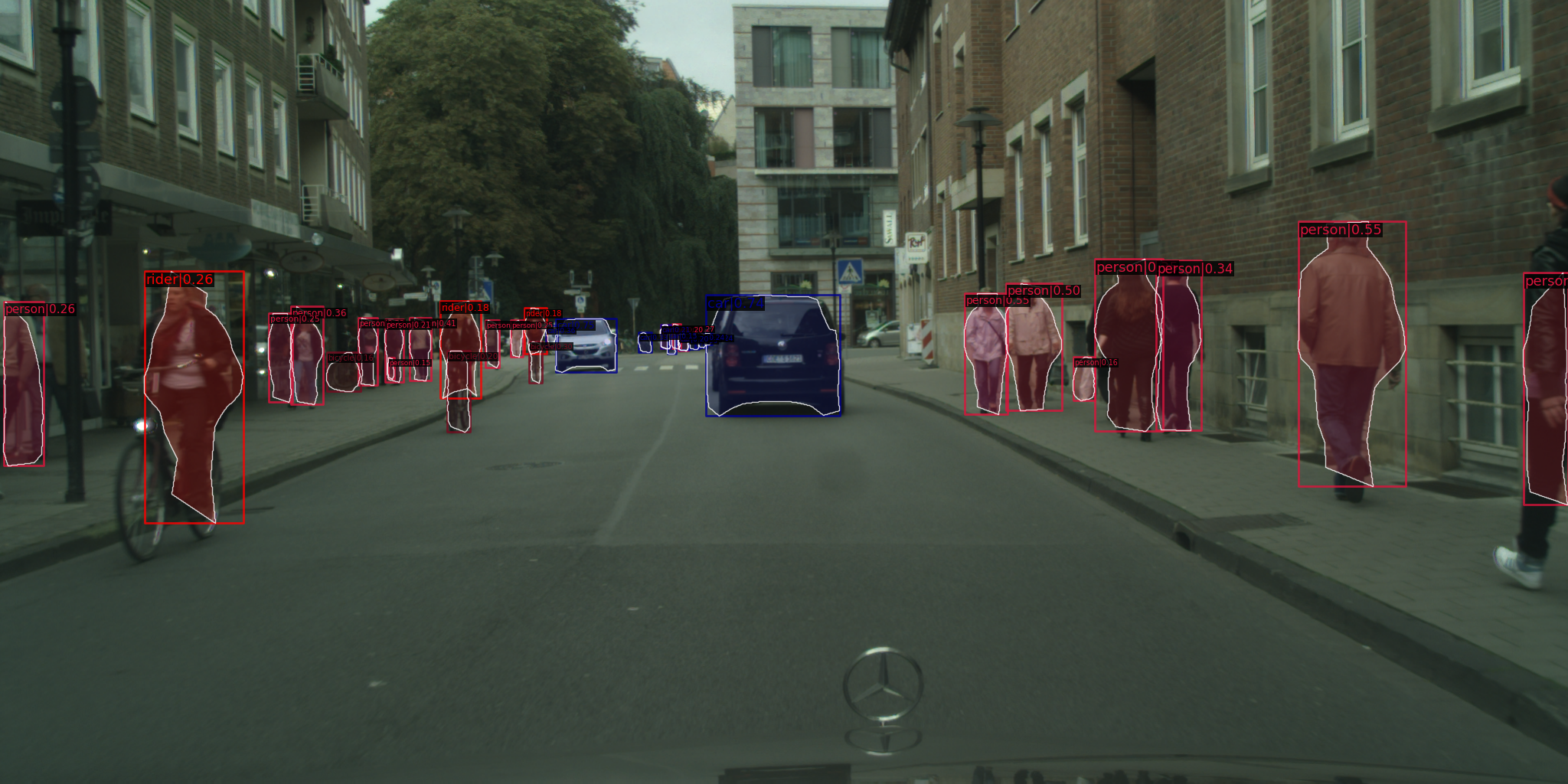}
    \end{subfigure}
    \caption{Qualitative visualizations of DPPD results on Cityscapes.}
    \label{fig:cityscapes-vis}
\end{figure*}

To examine the generic effectiveness, we experiment DPPD on the public Cityscapes \cite{cordts2015cityscapes} dataset, which covers more dynamic objects like vehicles, pedestians, bicycles, \etc. We pick Poly-YOLO (an improved version of PolarMask and YOLOv3) as our main competitor. As the Poly-YOLO used DarkNet-53 as the backbone, we select Resnet50 as the backbone for our DPPD for a fair comparison. Input resolution is $1024 \times 2048$. we set the number of prediction vertices $k=12$, and the number of resampling rays $m=360$. Results of DPPD is based on the official Cityscapes evaluation metrics at the instance level, tested on the validation split. Results of other methods are borrowed from the Poly-YOLO paper \cite{hurtik2022poly}.

As shown in \cref{tab:cityscapes}, for the accuracy, DPPD surpasses the state-of-the-art polygon method Poly-YOLO, even using a smaller number of vertices predicted ($12$ vs. $24$). Qualitative visualization are shown in \cref{fig:cityscapes-vis}. For the speed, DPPD runs at 56.1 FPS, measured on a Tesla V100 GPU. Since it is hard to reproduce competitors latency from the exact same environment, we mark it with a * symbol to indicate reference purpose only.



Poly-YOLO is built as an extension of YOLOv3. Comparing with DPPD, there are many fundamental differences such as the backbone structure, target assignment mechanisms, loss functions, \etc. Considering the polygon head itself, due to the fact of circular grid splits, Poly-YOLO is still learned from approximated labels. Poly-YOLO is trained for boxes and polygons jointly, and it is claimed benefited from the auxiliary task learning. However, it predicts the object center for both box and polygon, which is not guaranteed well aligned (see {\cref{fig:l-shape-mask}}). On the other hand, DPPD is a polygon-only detector. The resampling process allows DPPD to learn from the real polygon shape without any approximation, the angle decoding method enables its capability to fit for both sparse and dense vertices, and the object center encoded as the geometry centroid.


\begin{table}\small
  \caption{Benchmark on Cityscapes. Poly-YOLO predicts $24$ vertices, whereas DPPD predicts $12$ vertices. YOLOv3 and Poly-YOLO use DarkNet53 backbone, whereas MaskRCNN and DPPD use ResNet50 backbone.}
  \centering
  \begin{tabular}{c c c | c c c}
    \toprule
    Method 							& Detector	& $AP$		& $AP_{50}$	& FPS  \\
    \midrule
    YOLOv3 \cite{redmon2018yolov3}	& Box		& 10.6		& 26.6		& 26.3 \\
    MaskRCNN \cite{he2017mask}		& Mask		& 16.4		& 31.8		& 6.2  \\
    \midrule
    Poly-YOLO \cite{hurtik2022poly} (24) & Polygon	& 8.7		& 24.0		& 21.9 \\
    DPPD (12)	                    & Polygon	& 11.96		& 27.31		& 56.1* \\
    \bottomrule
  \end{tabular}
  \label{tab:cityscapes}
\end{table}

\subsection{Ablation Studies}

Admittedly, the optimization of model structure, data augmentation, and other fantastic training strategies could further improve the overall performance. But in this subsection, we focus on the key components involved in the polygon detector regression head. Specifically, we discuss the design choices of polygon origin finding methods, angle decoding methods, number of prediction vertices, number of resampling rays.

\begin{figure}[t]
  \centering
  \includegraphics[width=0.65\linewidth]{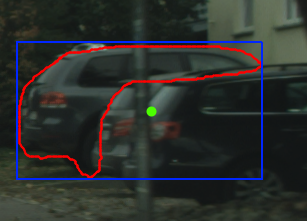}
  \caption{Example of polygon polar origin finding. If the vehicle object is partially occluded, the bounding box center (green) is outside of the polygon shape boundary (red).}
  \label{fig:l-shape-mask}
\end{figure}

\textbf{Polar Origin Finding.}
The polygon polar origin holds two regression targets. To find the groundtruth, we considered three methods: (i) the mean of all vertices; (ii) the bounding box center that covers the polygon; (iii) the polygon shape geometry centroid. However, (i) and (ii) are unable to guarantee that the polar origin is located within the polygon boundary (see \cref{fig:l-shape-mask}), which voids the model effectiveness. Therefore, (iii) geometry centroid is the only choice. We run experiments on a subset of crosswalk objects and compared the three methods, shown as Exp. 1-3 in \cref{tab:ablation}.

\textbf{Angle Decoding.}
The flexible angle prediction is one major distinction of DPPD. Comparing against PolarMask \cite{xie2020polarmask}, this is the second degree of freedom of each vertex; comparing against Poly-YOLO \cite{hurtik2022poly}, it breaks the angle sector constraint. Using the Cityscapes data, we experiment two angle decoding methods: (i) predict an offset within each angle bin; (ii) the angle cumulative summation. According to Exp. 4-5 in \cref{tab:ablation}, (ii) obtains higher accuracy than (i).

\textbf{Predictable Vertices and Resampling Rays.}
The number of predictable vertices and resampling rays are important hyperparameters to construct the DPPD head. We experiment six combinations of vertices and rays on Cityscapes. Results are listed in \cref{tab:ablation}. Increasing the number of rays (Exp.6 - 8) results in the accuracy improvement, this is expected since we have denser representation of the polygon shape. On the other hand, increasing the number of vertices (Exp.8 - 10) is less effective. Keep increasing it will lead to excessive model complexity and hence lower generalization capability. But this is a good indicator that DPPD is able to achieve promising results even with less regression outputs.

\begin{table*}\footnotesize
  \caption{Ablation studies for DPPD design choices.}
  \centering
  \begin{tabular}{c | c | c c c c | c c c | c c}
    \toprule
    Exp. & Dataset & Origin Finding & Angle Decoder & \# of Vertices & \# of Rays & $P$ & $R$ & $F1$ & $AP$ & $AP_{50}$ \\
    \midrule
    1	& Crosswalk 	& box center		& bin offsets	& 36		& 360	& 69.19	& 50.22	& 58.20	& - 		& -	\\
    2	& Crosswalk 	& vertices mean	& bin offsets	& 36		& 360	& 50.13	& 49.89	& 50.01	& - 		& -	\\
    3	& Crosswalk 	& geo centroid	& bin offsets	& 36		& 360	& 72.02	& 54.54	& 62.08	& - 		& -	\\
    \midrule
    4	& Cityscapes & geo centroid	& bin offsets	& 18		& 360	& -		& -		& -		& 11.10	& 25.80 \\
    5	& Cityscapes & geo centroid	& cumsum	& 18		& 360	& -		& -		& -		& 12.01 	& 27.45	\\
    \midrule
    6	& Cityscapes & geo centroid	& cumsum	& 12		& 120	& -		& -		& -		& 10.38	& 24.17 \\
    7	& Cityscapes & geo centroid	& cumsum	& 12		& 180	& -		& -		& -		& 11.54	& 27.29	\\
    8	& Cityscapes & geo centroid	& cumsum	& 12		& 360	& -		& -		& -		& 11.96	& 27.31	\\
    9	& Cityscapes & geo centroid	& cumsum	& 18		& 360	& -		& -		& -		& 12.01	& 27.45 \\
    10	& Cityscapes & geo centroid	& cumsum	& 36		& 360	& -		& -		& -		& 11.67	& 26.83 \\
    \bottomrule
  \end{tabular}
  \label{tab:ablation}
\end{table*}

\label{sec:experiment}

\section{Conclusion}
In this paper, we present DPPD, a deformable polygon detector that stands intermediately between object detection and instance segmentation. The polygon detector is able to describe precise object shape information, while retaining fast runtime inference speed. Our polygon detection method is able to predict object shapes with high accuacy without a need to use an excessively large number of vertices as done in the previous methods. This is possible due to our novel polygon training strategy, in which both ground truths and predictions (not necessarily having the same number of vertices) are up-sampled so that both have the same number of raypoints. The upsampling (resampling) is efficient and differentiable, which is required for training.  
From the experiments using both in-house autonomous driving dataset and public dataset, DPPD outperforms PolarMask and Poly-YOLO in terms of both accuracy and speed for many detection tasks such as road sign, crosswalk, car, pedestrian.

Although DPPD is designed as a 2D polygon detector, it is also applicable in 3D perception. Many state-of-the-art 3D detectors adopt BEV \cite{li2022bevformer} or rangeview \cite{fan2021rangedet} representations for runtime efficiency. We could place DPPD on top of any detection architecture for complex shape understanding.

\label{sec:conclusion}

{\small
\bibliographystyle{ieee_fullname}
\bibliography{11_references}
}


\end{document}


\title{\paperTitle \\ Supplemental Material}
\author{\authorBlock}
\maketitle

\appendix
\section{Appendix}

\subsection{Resampling Raypoints Calculation}

\textbf{Triangle Approach.}
In \cref{fig:5-a}, let $A$ and $B$ be two adjacent vertices and $O$ be the origin. The task is to find the intersection point $P$ between segment $AB$ and ray $\vec{OR}$. Deriving from the triangle similarity, we compute a ratio $r$:
\begin{equation}
  \begin{aligned}
  	r = \dfrac{|AP|}{|BP|} = \dfrac{|AC|}{|BD|} = \dfrac{|OA|\sin(\alpha)}{|OB|\sin(\beta)}
  \end{aligned}
  \label{eq:6}
\end{equation}
The ratio $r$ represents the position of $P$ within the segment $AB$. Then the point $P$ coordinates are calculated as:
\begin{equation}
  \begin{aligned}
  	(P_{x}, P_{y}) = (\dfrac{A_{x} + rB_{x}}{1+r}, \dfrac{A_{y} + rB_{y}}{1+r})
  \end{aligned}
  \label{eq:7}
\end{equation}

The coordinates of $O$, $A$, $B$ and their segment lengths are known from the decoding outputs. $\vec{OR}$ is one of the $M$ equal-spaced resampling rays. If we could determine which ray locates between $\vec{OA}$ and $\vec{OB}$, then the delta angles $\alpha$ and $\beta$ are also known. The problem is now transformed to finding two neighbor vertices of each ray, which could be solved by searching the ray angle within a sorted list of vertices angles.


\textbf{Vector Approach.} 
In \cref{fig:5-b}, let $\vec{u_{1}} = r - o$ be the unit vector along ray $\vec{OR}$, and $\vec{u_{2}} = b - a$ be the unit vector from $A$ to $B$. The intersection is to solve $t_{1}$ and $t_{2}$ from the equation:
\begin{equation}
  \begin{aligned}
	\begin{cases}
  	  f(t_{1}) = o + \vec{u_{1}} * t_{1}, t_{1} \in [0, \infty) \\
  	  f(t_{2}) = a + \vec{u_{2}} * t_{2}, t_{2} \in [0, 1]
  	\end{cases}
  \end{aligned}
  \label{eq:8}
\end{equation}
We constrain $t_{1} \in [0, \infty)$ to ensure the intersection P is along the positive direction of the ray, and $t_{2} \in [0, 1]$ to ensure P is at the segment between vertices. The mathematic solution is:
\begin{equation}
  \begin{aligned}
	\begin{cases}
  	  t_{1} = \dfrac{|\vec{v_{2}} \times \vec{v_{1}}|}{\vec{v_{2}} \cdot \vec{v_{3}}} \\
  	  t_{2} = \dfrac{\vec{v_{1}} \cdot \vec{v_{3}}}{\vec{v_{2}} \cdot \vec{v_{3}}}
  	\end{cases}
  \end{aligned}
  \label{eq:9}
\end{equation}
where $\vec{v_{1}} = o - a$ is the vector from $A$ to $O$; $\vec{v_{2}} = \vec{u_{2}} = b - a$ is the vector from $A$ to $B$; $\vec{v_{3}}$ is the vector perpendicular to ray vector $\vec{u_{1}}$. Suppose $\vec{u_{1}}$ is decomposed into $[u_{1x}, u_{1y}]$, then $\vec{v_{3}} = [-u_{1y}, u_{1x}]$. The operator $\cdot$ is the dot product and $\times$ is the cross product.


\label{sec:appendix-1}

{\small
\bibliographystyle{ieee_fullname}
\bibliography{11_references}
}